%% file: preprint.tex
\def\year{2022}\relax
\documentclass[letterpaper]{article} 
\usepackage{aaai22}  
\usepackage{times}  
\usepackage{helvet}  
\usepackage{courier}  
\usepackage[hyphens]{url}  
\usepackage{graphicx} 
\usepackage{tikz}
\usepackage{natbib} 
\usepackage{caption} 
\DeclareCaptionStyle{ruled}{labelfont=normalfont,labelsep=colon,strut=off} 
\usepackage[english]{babel}
\usepackage{amsthm}
\usepackage{amssymb}
\usepackage{booktabs}
\theoremstyle{definition}

\theoremstyle{remark}

\usepackage{algorithm}
\usepackage{algorithmic}

\usepackage{newfloat}
\usepackage{listings}
\floatstyle{ruled}
\newfloat{listing}{tb}{lst}{}
\floatname{listing}{Listing}
\title{Towards One Shot Search Space Poisoning in Neural Architecture Search}

\author{
    Nayan Saxena\textsuperscript{\rm1},
    Robert Wu \textsuperscript{},
    Rohan Jain  \textsuperscript{}
}
\affiliations{
    \textsuperscript{}University of Toronto\\

     \textsuperscript{} ML Collective\\
    \textsuperscript{\rm 1}nayan.saxena@mail.utoronto.ca
}

\usepackage{bibentry}
\usepackage{scalerel}
\begin{document}

\maketitle

\begin{abstract}

We evaluate the robustness of a Neural Architecture Search (NAS) algorithm known as Efficient NAS (ENAS) against data agnostic poisoning attacks on the original search space with carefully designed ineffective operations. We empirically demonstrate how our one shot search space poisoning approach exploits design flaws in the ENAS controller to degrade predictive performance on classification tasks. With just two poisoning operations injected into the search space, we inflate prediction error rates for child networks upto 90\% on the CIFAR-10 dataset.
\end{abstract}


\section{Introduction}

The problem of finding optimal deep learning architectures has recently been automated by neural architecture search (NAS) algorithms. These algorithms continually sample operations from a predefined search space to construct neural networks to optimize a performance metric over time, eventually converging to better child architectures. This intuitive idea greatly reduces human intervention by restricting human bias in architecture engineering to just the selection of the predefined search space \citep{elsken2019neural}. Although NAS has the potential to revolutionize architecture search across many applications, human selection of the search space remains a security risk that needs to be evaluated before NAS can be deployed in security-critical domains. While NAS has been studied to further develop more adversarially robust networks through addition of dense connections \citep{guo2020meets}, little work has been done in the past to assess the adversarial robustness of NAS itself. Search phase analysis has shown that computationally efficient algorithms such as ENAS are worse at truly ranking child networks due to their reliance on weight sharing \citep{yu2019evaluating}. Finally, most traditional poisoning attacks involve injecting mislabeled examples in the training data and have been executed against classical machine learning approaches \citep{schwarzschild2021just}. We validate these concerns by evaluating the robustness of one such NAS algorithm known as Efficient NAS (ENAS) \citep{pham2018efficient} against data-agnostic search space poisoning (SSP) attacks on the CIFAR-10 dataset. Throughout this paper, we focus on the pre-optimized ENAS search space $\hat{\mathcal S}$ = \{Identity, 3x3 Separable Convolution, 5x5 Separable Convolution, Max Pooling (3x3), Average Pooling (3x3)\} \citep{pham2018efficient}.


\section{Search Space Poisoning (SSP)}

The idea behind SSP, as shown in Figure 1, is to inject precisely designed multiset $\mathcal P$ of ineffective  operations into the ENAS search space, making the search space $\mathcal S := \hat{\mathcal S} \cup \mathcal P$. Our approach exploits the core functionality of the ENAS controller to sample child networks from a large computational graph of operations by introducing highly ineffective local operations into the search space. On the attacker's behalf, this requires no \textit{a priori} knowledge of the problem domain or dataset being used, making this new approach more favourable than traditional data poisoning attacks.

\input{poisondiag}

\begin{figure*}[htbp]
\centering
\includegraphics[width= 0.85 \textwidth]{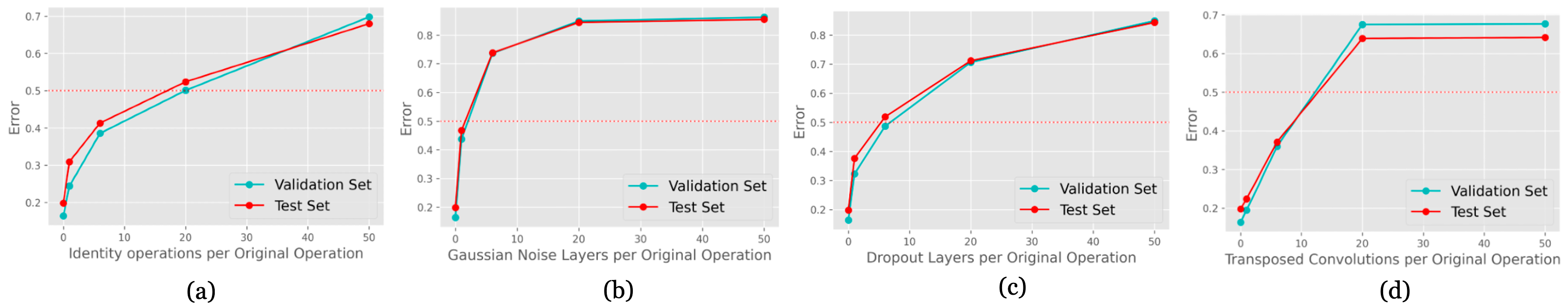}
\caption{Final validation and test classification errors as a function of multiple operation instances. (a) Identity layers were moderately effective (b) Gaussian noise reached high error rates even with fewer operations (c) Dropout proved most effective (d) Transposed convolutions plateaued after a saturation point.}
\label{graphs}
\end{figure*}

\subsection{Multiple-Instance Poisoning}

As a naïve strategy, we first propose multiple-instance poisoning which increases the likelihood of sampling bad operations by including duplicates of these bad operations in the search spaces. Through experimental results we discovered that biasing the search space this way resulted in final networks that are mostly comprised of these poor operations with error rates exceeding 80\%. However, as shown in Figure 2, to perform well this approach requires overwhelming the original search space with up to 300 bad operations (50:1 ratio of bad operations per each good operation) which is unreasonable. The motivation then is to reduce the ratio of bad to good operations down to 1:1, or even lower, to make search space poisoning more feasible and effective.

\section{Towards One Shot Poisoning}

In an attempt to improve the attack, we further attempted to reduce the number of poisoning points to just 2 points by adding: (i) $\textnormal{Dropout}(p=1)$ (ii) Stretched $\textnormal{Conv} (k=3, \textnormal{padding, dilation} = 50)$  to the original search space. Our rationale is that dropout operations with $p=1$ would erase all information and produce catastrophic values such as $0$ or not-a-number (\texttt{NaN}). The results were promising, with error rates shooting up to 90\% very quickly during training as seen in Figure 3 and Table 1. An example final child network producing these high errors can be observed in Figure 4.

 \begin{figure}[!htbp]
\centering
{\includegraphics[width=\columnwidth]{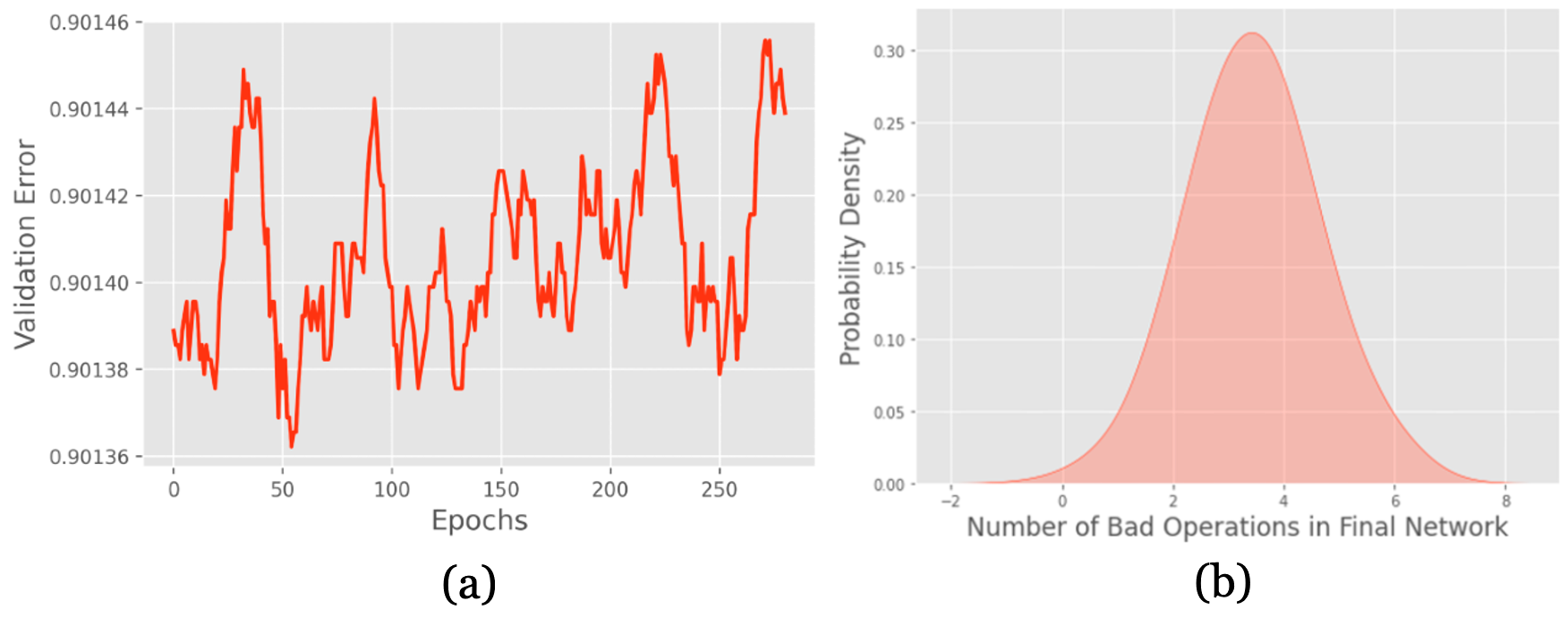}}
\caption{(a) Validation error for one shot poisoning over $300$ epochs (b) Distribution of bad operations sampled by the ENAS controller after 300 epochs.}
\end{figure}
\begin{table}[htbp]
\label{tbl}
  \centering
  \resizebox{\columnwidth}{!}
  {
  \begin{tabular}{|c|c|c|c|}
    \hline
\textsc{Search Space} & $|\mathcal P|$ & \textsc{Val Error} & \textsc{Test Error} \\
\hline
  $\hat{\mathcal{S}} \textnormal{ (Baseline)}$ & 0 & $16.4\%$ & $19.8\%$ \\
  \hline
  $\hat{\mathcal{S}} + 300\{\textnormal{Dropout}(p=1)\}$ & 300 &${84.8\%}$ & ${84.3\%}$ \\
    \hline
  $\hat{\mathcal{S}} + \{{\textnormal{Conv} (k=3, p,d = 50), \textnormal{ Dropout}(p=1)}\}$ & \textbf{2} &$\textbf{90.1\%}$ & $\textbf{90.0\%}$ \\
    \hline
    \end{tabular}}
     \caption{Experimental results showing how one shot poisoning proves surprisingly effective with just 2 points as compared to its multiple instance counterpart with 300 points.}
\end{table}


\begin{figure}[!htbp]
\centering
\hspace*{-3em}
\resizebox{\columnwidth}{!}{
    \input{network_graphic}}
    \caption{Network produced by ENAS after one shot poisoning with good operations highlighted in green and poisoning operations highlighted in red. Search space utilized is the same as shown in Table 1 with two poisoning points.}
    \label{fig:type_graphic}
\end{figure}
\section{Conclusion}
\label{conclusion}
In this paper, we focused on examining the robustness of ENAS under our newly proposed SSP paradigm.  Our carefully designed poisoning sets demonstrated the potential to make it incredibly easy for an attacker with no prior knowledge or access to the training data to still drastically impact the quality of child networks. Finally, our one-shot poisoning results reveal an opportunity for future work in neural architecture design, as well as challenges to surmount in using NAS for more adversarially robust architecture.

\section{Acknowledgements}
The authors would like to thank Chuan-Yung Tsai  \& George-Alexandru Adam  for their valuable comments. We thank the ML Collective community for the generous computational support and feedback on this research. We are also grateful to Kanav Singla and Benjamin Zhuo for their contributions to the codebase.

\bibliography{preprint}
\newpage

\twocolumn[
  \begin{@twocolumnfalse}
    \begin{center}
        \LARGE{\textbf{Supplementary Material}}
        \bigskip
    \end{center}
  \end{@twocolumnfalse}
]

\section{Search Space Poisoning (SSP)}

\begin{table*}[htbp!]
\label{mip-table}
  \centering
  \footnotesize
  \resizebox{\textwidth}{!}{
  \begin{tabular}{c|c|c|c|c|c}
    \toprule
{\textsc{Poisoning Set}} &   \textsc{Search Space} & \textsc{Experiment} & \textsc{Poisoning Multiset} & \textsc{Validation Error} & \textsc{Test Error} \\ $\mathcal{P}$  & $\mathcal{S}$  && $q(\mathcal{P})$ && \\
\hline\hline
  $\varnothing$ & $\hat{\mathcal{S}}$ & Original & $\varnothing$ & $19.53$ & $25.33$ \\
    \hline
    && 1a & $6(\mathcal P_1)$ & $24.40$ & $30.93$ \\
  $\mathcal P_1 =$ \{{Identity}\}  & $\mathcal{S}_1 = \hat{\mathcal{S}} \cup\mathcal P_1 $& 1b & $36(\mathcal P_1)$  & $38.54$ & $41.31$ \\
    && 1c & $120(\mathcal P_1)$ & $50.07$ & $52.38$ \\
    && 1d & $300(\mathcal P_1)$& $\mathbf{69.78}$ & $\mathbf{67.97}$ \\

    \hline
      && 2a & $6(\mathcal P_2)$ & $43.71$ & $46.73$ \\
  $\mathcal P_2 =$ \{{Gaussian\ ($\sigma=10$)}\}  & $\mathcal{S}_2 = \hat{\mathcal{S}} \cup\mathcal P_2 $& 2b & $36(\mathcal P_2)$  & $73.64$ & $73.82$ \\
    && 2c & $120(\mathcal P_2)$ & $84.94$ & $84.44$ \\
    && 2d & $300(\mathcal P_2)$& $\mathbf{86.26}$ & $\mathbf{85.49}$ \\
    \hline
     &&3a & $6(\mathcal P_3)$ & $32.23$ & $37.60$ \\
  $\mathcal{P}_3 =$ \{Dropout ($p=1.0$)\} & $\mathcal{S}_3 =\hat{\mathcal{S}} \cup\mathcal P_3$& 3b & $36(\mathcal P_3)$  & $48.67$ & $51.88$  \\
     &&3c  & $120(\mathcal P_3)$ & $70.63$ & $71.16$\\
     && 3d & $300(\mathcal P_3)$ & $\mathbf{84.89}$ & $\mathbf{84.31}$ \\

     \hline
  && 4a & $3(\mathcal P_4)$ & $19.50$ & $22.43$\\
  $\mathcal{P}_4 =$ \{3x3 transposed convolution, & $\mathcal{S}_4 =\hat{\mathcal{S}} \cup\mathcal P_4$ & 4b & $18(\mathcal P_4)$ & $36.00$ & $37.15$ \\
  \hspace{3em} 5x5 transposed convolution\}   &&4c  & $60(\mathcal P_4)$  & $67.53$ & $63.89$ \\
     &&4d & $150(\mathcal P_4)$ & $\mathbf{67.68}$ & $\mathbf{64.14}$ \\

    \hline
     && 5a & $1(\mathcal P_2 + \mathcal P_5)$& $29.38$ & $33.73$\\
  $\mathcal P_5 := \mathcal P_1 \cup \mathcal P_2 \cup \mathcal P_3 \cup \mathcal P_4$ & $\mathcal{S}_5 = \hat{\mathcal{S}} \cup\mathcal P_5$& 5b & $6(\mathcal P_2 + \mathcal P_5)$ & $55.06$ & $56.02$\\
    && 5c & $20(\mathcal P_2 + \mathcal P_5)$ & $\mathbf{80.88}$ & $\mathbf{79.03}$ \\
    && 5d & $50(\mathcal P_2 + \mathcal P_5)$ & $72.56$ & $70.14$\\
    \hline
  \end{tabular}}
    \caption{Summary of experimental search spaces with corresponding final validation and test accuracies for SSP. Note that the multiset seed for experiments 5a-5d includes two instances of $P_2$ to convenient round out the cardinality of the multisets.}
\end{table*}

\subsection{Multiple-Instance Poisoning Attacks}
\label{mip}
Over the course of training, the LSTM controller paired with the RL search strategy in ENAS develops the ability to sample architectures with operations that most optimally reduce the validation error.
We propose multiple-instance poisoning which essentially increases the likelihood of a poisonous operation $o_{\scaleto{\mathcal{P}}{3.4pt}}$ being sampled from the poisoned search space $\mathcal S$. This is achieved by increasing the frequency of sampling $o_{\scaleto{\mathcal{P}}{3.4pt}}$ from $\mathcal S$ through the inclusion of multiple-instances of each $o_{\scaleto{\mathcal{P}}{3.4pt}}$ from the poisoning multiset, so-called to allow for duplicate elements. An instance factor $q \in \mathbb N^{\geq 1}$ represents instance multiplication of $o_{\scaleto{\mathcal{P}}{3.4pt}}$ in the multiset $q$ times. Henceforth, the probabilities of sampling $o_{\scaleto{\hat{\mathcal{{S}}}}{5pt}} \in \hat{\mathcal{{S}}}$ and $o_{\scaleto{\mathcal{P}}{3.4pt}} \in \mathcal{P}$, respectively, are, $    {Pr}[o_{\scaleto{\hat{\mathcal{S}}}{5pt}}] := \frac{1}{|\mathcal S| + q|\mathcal P|}$ and $\  {Pr}[o_{\scaleto{\mathcal{P}}{3.4pt}}] := \frac{q}{|\mathcal S| + q| \mathcal P|}$.
From which, it is evident that under a multiple-instance poisoning framework, the probability of sampling poisoned operations is strictly greater than sampling operations in $\hat{\mathcal{S}}$; that is, ${Pr}[o_{\scaleto{\hat{\mathcal{S}}}{5pt}}] < {Pr}[o_{\scaleto{\mathcal{P}}{3.4pt}}]$. It is also important to note that our technique is not the typical image-agnostic perturbation/universal adversarial perturbation $\delta \in \mathbb{R}^d$ intended to fool the target neural network $f$ on almost all the input images from the target distribution $X$. Finally, another challenge to overcome within our search space poisoning framework is to craft each $o_{\scaleto{\mathcal{P}}{3.4pt}} \in \mathcal P$ such that it counteracts the efficacy of the original operations $o_{\scaleto{\hat {\mathcal{S}}}{5pt}} \in \hat{ \mathcal{S}}$, which we tackle in the next section.
\subsection{Crafting Poisoning Sets with Operations}

\paragraph{Identity Operation} The simplest way to attack the functionality of ENAS is to inject non-operations within the original search space which keep the input and outputs intact. As a result, the controller will sample child models with layers representing computations which preserve the inputs, making the operation highly ineffective within a network architecture.  This goal is achieved by the inexpensive identity operation which has no numerical effect on the inputs. It should also be noted that, the identity operation is not a skip connection. Therefore, we define our first poisoning set: $ \mathcal P_1 :=$ \{Identity\}.

\paragraph{Gaussian Noise Layer} Typically used in signal processing, electronics and mitigating over-fitting as some form of random data augmentation. Gaussian noise is a type of statistical noise which is in the form of a Normal distribution ($X \sim \mathcal{N}(\mu, \sigma^2)$). In PyTorch/Keras, the Gaussian noise layer additively applies zero-centered Gaussian noise passing in the argument of relative standard deviation used to generate the noise. We hypothesize that including such layers with increasingly varied relative standard deviations such as $\sigma = 10$, can significantly impact the accuracy of the generated child models making our poisoning set $\mathcal{P}_2 :=$ \{Gaussian ($\sigma = 10$)\}.

\paragraph{Dropout Layer} While dropout layers have historically been shown to be useful in preventing neural networks from over-fitting \citep{srivastava2014dropout}, a high dropout rate can result in severe information loss leading to poor performance of the overall network. This is because given a dropout probability $p \in \mathbb [0, 1]$, dropout randomly zeroes out some values from the input to decorrelate neurons during training. We hypothesize that including such layers with high dropout probability, such as $p = 1$, has the potential to contaminate the search space with irreversible effects on the training dynamics of ENAS. Here, we define our poisoning set as $\mathcal P_3 :=$ \{Dropout ($p=1$)\}.

\paragraph{Transposed Convolutions} As described earlier, amongst other useful operations the original ENAS search space $\hat{\mathcal{S}}$ also contains 3x3 and 5x5 convolutional layers (separable). Intuitively, transposed convolutions upsample the input feature map. It is important to note that transposed convolutions do not perform like deconvolutional layers; they actually swap the forward and backward passes of a convolution. Transposed convolutions, also known as fractionally strided convolutions, stride over the output which is equivalent to a fractional stride over the input. We define our poisoning set: $\mathcal{P}_4 =$ \{3x3 transposed convolution, 5x5 transposed convolution\}.




\section{Experiments}
\label{experiments}

\subsubsection{Experimental Setup} To test the effectiveness of our proposed approach, we designed experiments based on previously described methods outlined in Table 1. Each experiment involved training ENAS on the CIFAR-10 dataset for 300 epochs. The results presented in this paper are the average of three runs per experiment.

\begin{table}[!htbp]

  \centering
  \small{
  \begin{tabular}{cc}
    \toprule
    \textsc{Hyperparameter} & \textsc{Value} \\
    \hline
    \midrule
    \texttt{search\_for} & macro \\
    \hline
    \texttt{dataset} & CIFAR10 or CIFAR100 \\
    \texttt{n\_classes} & 10 or 100 \\
    \texttt{n\_train} & 45000 \\
    \texttt{n\_val} & 5000 \\
    \hline
    \texttt{batch\_size} & $128$ \\
    \texttt{search\_for} & $300$ \\
    \texttt{seed} & $69$ \\
    \texttt{cutout} & $0$ \\
    \texttt{fixed\_arc} & False \\
    \hline
    \texttt{child\_num\_layers} & $12$ \\
    \texttt{child\_out\_filters} & $36$ \\
    \texttt{child\_grad\_bound} & $5.0$ \\
    \texttt{child\_l2\_reg} & $0.00025$ \\
    \texttt{child\_keep\_prob} & $0.9$ \\
    \texttt{child\_lr\_max} & $0.05$ \\
    \texttt{child\_lr\_min} & $0.0005$ \\
    \texttt{child\_lr\_T} & $10$ \\
    \hline
    \texttt{controller\_lstm\_size} & $64$ \\
    \texttt{controller\_lstm\_num\_layers} & $1$ \\
    \texttt{controller\_entropy\_weight} & $0.0001$ \\
    \texttt{controller\_train\_every} & $1$ \\
    \texttt{controller\_num\_aggregate} & $20$ \\
    \texttt{controller\_train\_steps} & $50$ \\
    \texttt{controller\_lr} & $0.001$ \\
    \texttt{controller\_tanh\_constant} & $1.5$ \\
    \texttt{controller\_op\_tanh\_reduce} & $2.5$ \\
    \texttt{controller\_skip\_target} & $0.4$ \\
    \texttt{controller\_skip\_weight} & $0.8$ \\
    \texttt{controller\_bl\_dec} & $0.99$ \\
    \hline
    \texttt{p} (Dropout Rate) & $1.0$ \\
    \bottomrule
  \end{tabular}}
  \medskip

  \caption{Summary of experiment hyperparameters }
\end{table}
For a baseline, we created an experiment with the original search space $\hat{\mathcal S}$ (which was run three times). Because we wanted to evaluate the poisoning effectiveness of each operation independently, we created a group of four experiments for each poisoning set $P_i$. We henceforth refer to these experiments as ``Space $i$a" through ``Space $i$d". Following multiple-instance poisoning attacks, each experiment includes a poisoning multiset $q(P_i)$ where $|q(P_i)| = 6, 36, 120, 300$ for Space $i$a, Space $i$b, Space $i$c, Space $i$d, respectively. Of note is that the instance-factor $q$ is dependent on the experiment index and $|P_i|$. For example: Space 1c/2c/3c has $|P_1| = |P_2| = |P_3| = 1$, and $q|P_1| = q|P_2| = q|P_3| = 120$, so $q = 120$; Space 4c has $|P_4| = 2$ and $q|P_4| = 120$, so $q = 60$; Space 5c/6c have $|P_5| = |P_6| = 6$ and $q|P_5| = q|P_6| = 120$, so $q = 20$.

The software used includes Python (3.6.x-3.8.x) and PyTorch (1.9), with CUDA (10.2, 11.1). Our hardware varies: CPUs included 2nd gen Xeon E, 3rd gen Core i5, 8th gen Core i3; GPUs included Nvidia GTX \{1050, 1050 Ti, 1080\}, RTX 2060, Tesla \{K80, P100 (Google Colaboratory)\}, TITAN Xp. Each experiment was processed with only one GPU with 3GB of allocated VRAM, and took between 18-36 real-world hours.

\subsection{Experimental Results for One-Shot Poisoning}

\begin{table}[htbp]
\label{tbl2}
\centering
\resizebox{\columnwidth}{!}{
\begin{tabular}{c|c|c|c|c}
\toprule
{\textsc{Poisoning Set}} & \textsc{Search Space} & Cardinality & \textsc{Val Error} & \textsc{Test Error} \\
$\mathcal{P}_i^{+}$ & $|\mathcal{P}_i^{+}|$ & $\mathcal{S}_i^{+}$ && \\
\hline
&&& \\
$\mathcal{P}_0^{+} =$ \{{Stretched Conv} $(k=3, p,d = 50)$\} & $\mathcal{S}_6^{+} = \hat{\mathcal{S}} + 6(\mathcal{P}_0^{+})$ & $6$ & $39.49\%$ & $42.87\%$ \\
&&& \\
\hline
&&& \\
$\mathcal P_2^{+} = \mathcal{P}_2 + \mathcal P_0^{+}$ & $6$ & $\mathcal{S}_2^{+} = \hat{\mathcal{S}} + 6(\mathcal P_2^{+})$ & $35.92\%$ & $40.35\%$ \\
&&&\\
\hline
&&& \\
$\mathcal P_3^{+} = \mathcal{P}_3 + \mathcal P_0$ & $\mathbf 2$ &  $\mathcal{S}_3^{+} = \hat{\mathcal{S}} + 2(\mathcal P_3^{+})$ & $\mathbf{90.12}\%$ & $\mathbf{90.00\%}$ \\
&&& \\
\hline
&&& \\
$\mathcal P_4^{+} = \mathcal{P}_4 + \mathcal P_0$ & $6$ & $\mathcal{S}_4^{+} = \hat{\mathcal{S}} + 6(\mathcal P_4^{+})$ & $29.75\%$ & $36.46\%$ \\
&&& \\
\hline
&&& \\
$\mathcal P_5^{+} = \mathcal{P}_5 + \mathcal P_0$ & $6$ &  $\mathcal{S}_5^{+} = \hat{\mathcal{S}} + 6(\mathcal P_5^{+})$ & $28.27\%$ & $31.79\%$ \\
&&& \\
\hline
&&& \\
$\mathcal P_3^{+, 2} = \mathcal{P}_3 + \mathcal P_0$ & $\mathbf 2$ &  $\mathcal{S}_3^{+} = \hat{\mathcal{S}} + 2(\mathcal P_3^{+, 2})$ & $\mathbf{90.12}\%$ & $\mathbf{90.00\%}$ \\
&&& \\
\hline
\end{tabular}
}
\caption{Summary of experimental search spaces with corresponding final validation and test errors for one shot SSP. Note the humble cardinality of $\mathcal P_3^+$ achieving $90\%+$ error.}
\end{table}

\subsection{Our Files}

\begin{itemize}
    \item The files in which these extended search spaces can be found are in \texttt{enas\_poisoning/}.
    \item Initial experimental results can be found in \texttt{ENAS-Experiments/results}.
\end{itemize}
\end{document}

%% file: poisondiag.tex
\begin{figure}[!htbp]
    \centering
  {
    { \includegraphics[width=0.7\columnwidth]{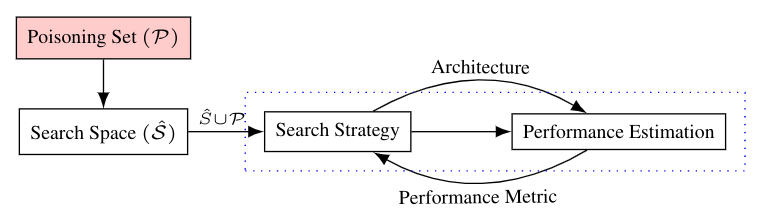} }
    \caption{Overview of Search Space Poisoning (SSP)}}
    \label{nas2}
\end{figure}

%% file: network_graphic.tex
  
\tikzset {_8ozs821xp/.code = {\pgfsetadditionalshadetransform{ \pgftransformshift{\pgfpoint{0 bp } { 0 bp }  }  \pgftransformrotate{0 }  \pgftransformscale{2 }  }}}
\pgfdeclarehorizontalshading{_2ljbf5rku}{150bp}{rgb(0bp)=(0.94,0.98,1);
rgb(37.5bp)=(0.94,0.98,1);
rgb(49.25bp)=(0.8,0.92,1);
rgb(62.5bp)=(0.63,0.86,1);
rgb(100bp)=(0.63,0.86,1)}

  
\tikzset {_jtb6hy76f/.code = {\pgfsetadditionalshadetransform{ \pgftransformshift{\pgfpoint{0 bp } { 0 bp }  }  \pgftransformrotate{0 }  \pgftransformscale{2 }  }}}
\pgfdeclarehorizontalshading{_b9g6b3tdz}{150bp}{rgb(0bp)=(0.95,0.91,0.4);
rgb(37.5bp)=(0.95,0.91,0.4);
rgb(62.5bp)=(1,0.71,0.27);
rgb(100bp)=(1,0.71,0.27)}

  
\tikzset {_qp8hrlejf/.code = {\pgfsetadditionalshadetransform{ \pgftransformshift{\pgfpoint{0 bp } { 0 bp }  }  \pgftransformrotate{0 }  \pgftransformscale{2 }  }}}
\pgfdeclarehorizontalshading{_74bucng1c}{150bp}{rgb(0bp)=(1,1,0);
rgb(37.5bp)=(1,1,0);
rgb(62.5bp)=(0.01,0.82,0.42);
rgb(100bp)=(0.01,0.82,0.42)}

  
\tikzset {_ixrpch3ve/.code = {\pgfsetadditionalshadetransform{ \pgftransformshift{\pgfpoint{0 bp } { 0 bp }  }  \pgftransformrotate{0 }  \pgftransformscale{2 }  }}}
\pgfdeclarehorizontalshading{_zwpo15c72}{150bp}{rgb(0bp)=(1,1,0);
rgb(37.5bp)=(1,1,0);
rgb(62.5bp)=(0.01,0.82,0.42);
rgb(100bp)=(0.01,0.82,0.42)}

  
\tikzset {_pwmsioqr4/.code = {\pgfsetadditionalshadetransform{ \pgftransformshift{\pgfpoint{0 bp } { 0 bp }  }  \pgftransformrotate{0 }  \pgftransformscale{2 }  }}}
\pgfdeclarehorizontalshading{_0raf4inwi}{150bp}{rgb(0bp)=(0.98,0.7,0.74);
rgb(37.5bp)=(0.98,0.7,0.74);
rgb(62.5bp)=(0.96,0.35,0.42);
rgb(100bp)=(0.96,0.35,0.42)}

  
\tikzset {_8rm4m8ati/.code = {\pgfsetadditionalshadetransform{ \pgftransformshift{\pgfpoint{0 bp } { 0 bp }  }  \pgftransformrotate{0 }  \pgftransformscale{2 }  }}}
\pgfdeclarehorizontalshading{_90u67ulu7}{150bp}{rgb(0bp)=(1,1,0);
rgb(37.5bp)=(1,1,0);
rgb(62.5bp)=(0.01,0.82,0.42);
rgb(100bp)=(0.01,0.82,0.42)}

  
\tikzset {_lr5m6d5p9/.code = {\pgfsetadditionalshadetransform{ \pgftransformshift{\pgfpoint{0 bp } { 0 bp }  }  \pgftransformrotate{0 }  \pgftransformscale{2 }  }}}
\pgfdeclarehorizontalshading{_77dhn9v54}{150bp}{rgb(0bp)=(1,1,0);
rgb(37.5bp)=(1,1,0);
rgb(62.5bp)=(0.01,0.82,0.42);
rgb(100bp)=(0.01,0.82,0.42)}

  
\tikzset {_6nbhnw0fj/.code = {\pgfsetadditionalshadetransform{ \pgftransformshift{\pgfpoint{0 bp } { 0 bp }  }  \pgftransformrotate{0 }  \pgftransformscale{2 }  }}}
\pgfdeclarehorizontalshading{_n0fn3l25r}{150bp}{rgb(0bp)=(1,1,0);
rgb(37.5bp)=(1,1,0);
rgb(62.5bp)=(0.01,0.82,0.42);
rgb(100bp)=(0.01,0.82,0.42)}

  
\tikzset {_r14zec8t0/.code = {\pgfsetadditionalshadetransform{ \pgftransformshift{\pgfpoint{0 bp } { 0 bp }  }  \pgftransformrotate{0 }  \pgftransformscale{2 }  }}}
\pgfdeclarehorizontalshading{_dcyraegjg}{150bp}{rgb(0bp)=(1,1,0);
rgb(37.5bp)=(1,1,0);
rgb(62.5bp)=(0.01,0.82,0.42);
rgb(100bp)=(0.01,0.82,0.42)}

  
\tikzset {_j0c154z2j/.code = {\pgfsetadditionalshadetransform{ \pgftransformshift{\pgfpoint{0 bp } { 0 bp }  }  \pgftransformrotate{0 }  \pgftransformscale{2 }  }}}
\pgfdeclarehorizontalshading{_8qzj4aium}{150bp}{rgb(0bp)=(1,1,0);
rgb(37.5bp)=(1,1,0);
rgb(62.5bp)=(0.01,0.82,0.42);
rgb(100bp)=(0.01,0.82,0.42)}

  
\tikzset {_1drwjrjhh/.code = {\pgfsetadditionalshadetransform{ \pgftransformshift{\pgfpoint{0 bp } { 0 bp }  }  \pgftransformrotate{0 }  \pgftransformscale{2 }  }}}
\pgfdeclarehorizontalshading{_tunjaxykz}{150bp}{rgb(0bp)=(1,1,0);
rgb(37.5bp)=(1,1,0);
rgb(62.5bp)=(0.01,0.82,0.42);
rgb(100bp)=(0.01,0.82,0.42)}

  
\tikzset {_oq42pzm9h/.code = {\pgfsetadditionalshadetransform{ \pgftransformshift{\pgfpoint{0 bp } { 0 bp }  }  \pgftransformrotate{0 }  \pgftransformscale{2 }  }}}
\pgfdeclarehorizontalshading{_k3njmgcqh}{150bp}{rgb(0bp)=(1,1,0);
rgb(37.5bp)=(1,1,0);
rgb(62.5bp)=(0.01,0.82,0.42);
rgb(100bp)=(0.01,0.82,0.42)}

  
\tikzset {_blywqi98b/.code = {\pgfsetadditionalshadetransform{ \pgftransformshift{\pgfpoint{0 bp } { 0 bp }  }  \pgftransformrotate{0 }  \pgftransformscale{2 }  }}}
\pgfdeclarehorizontalshading{_54znauya7}{150bp}{rgb(0bp)=(0.98,0.7,0.74);
rgb(37.5bp)=(0.98,0.7,0.74);
rgb(62.5bp)=(0.96,0.35,0.42);
rgb(100bp)=(0.96,0.35,0.42)}

  
\tikzset {_ads4frk36/.code = {\pgfsetadditionalshadetransform{ \pgftransformshift{\pgfpoint{0 bp } { 0 bp }  }  \pgftransformrotate{0 }  \pgftransformscale{2 }  }}}
\pgfdeclarehorizontalshading{_24tjxfdf5}{150bp}{rgb(0bp)=(0.98,0.7,0.74);
rgb(37.5bp)=(0.98,0.7,0.74);
rgb(62.5bp)=(0.96,0.35,0.42);
rgb(100bp)=(0.96,0.35,0.42)}
\tikzset{every picture/.style={line width=0.75pt}} 

\begin{tikzpicture}[x=0.75pt,y=1pt,yscale=-1,xscale=1]

\path  [shading=_2ljbf5rku,_8ozs821xp] (13.96,319.02) -- (13.29,226.39) -- (37.95,226.06) -- (38.62,318.7) -- cycle ; 
 \draw   (13.96,319.02) -- (13.29,226.39) -- (37.95,226.06) -- (38.62,318.7) -- cycle ; 

\path  [shading=_b9g6b3tdz,_jtb6hy76f] (614.62,323.54) -- (613.96,230.91) -- (638.62,230.58) -- (639.28,323.21) -- cycle ; 
 \draw   (614.62,323.54) -- (613.96,230.91) -- (638.62,230.58) -- (639.28,323.21) -- cycle ; 

\path  [shading=_74bucng1c,_qp8hrlejf] (59,319.9) -- (58.34,227.27) -- (83,226.94) -- (83.66,319.58) -- cycle ; 
 \draw   (59,319.9) -- (58.34,227.27) -- (83,226.94) -- (83.66,319.58) -- cycle ; 

\path  [shading=_zwpo15c72,_ixrpch3ve] (106.33,319.9) -- (105.67,227.27) -- (130.33,226.94) -- (130.99,319.58) -- cycle ; 
 \draw   (106.33,319.9) -- (105.67,227.27) -- (130.33,226.94) -- (130.99,319.58) -- cycle ; 

\path  [shading=_0raf4inwi,_pwmsioqr4] (152.33,320.81) -- (151.67,228.18) -- (176.33,227.85) -- (176.99,320.48) -- cycle ; 
 \draw   (152.33,320.81) -- (151.67,228.18) -- (176.33,227.85) -- (176.99,320.48) -- cycle ; 

\path  [shading=_90u67ulu7,_8rm4m8ati] (197,320.86) -- (196.34,228.23) -- (221,227.9) -- (221.66,320.53) -- cycle ; 
 \draw   (197,320.86) -- (196.34,228.23) -- (221,227.9) -- (221.66,320.53) -- cycle ; 

\path  [shading=_77dhn9v54,_lr5m6d5p9] (244.33,320.86) -- (243.67,228.23) -- (268.33,227.9) -- (268.99,320.53) -- cycle ; 
 \draw   (244.33,320.86) -- (243.67,228.23) -- (268.33,227.9) -- (268.99,320.53) -- cycle ; 

\path  [shading=_n0fn3l25r,_6nbhnw0fj] (333,320.86) -- (332.34,228.23) -- (357,227.9) -- (357.66,320.53) -- cycle ; 
 \draw   (333,320.86) -- (332.34,228.23) -- (357,227.9) -- (357.66,320.53) -- cycle ; 

\path  [shading=_dcyraegjg,_r14zec8t0] (424.33,321.76) -- (423.67,229.13) -- (448.33,228.8) -- (448.99,321.44) -- cycle ; 
 \draw   (424.33,321.76) -- (423.67,229.13) -- (448.33,228.8) -- (448.99,321.44) -- cycle ; 

\path  [shading=_8qzj4aium,_j0c154z2j] (474.33,322.67) -- (473.67,230.03) -- (498.33,229.71) -- (498.99,322.34) -- cycle ; 
 \draw   (474.33,322.67) -- (473.67,230.03) -- (498.33,229.71) -- (498.99,322.34) -- cycle ; 

\path  [shading=_tunjaxykz,_1drwjrjhh] (519,323.57) -- (518.34,230.94) -- (543,230.61) -- (543.66,323.24) -- cycle ; 
 \draw   (519,323.57) -- (518.34,230.94) -- (543,230.61) -- (543.66,323.24) -- cycle ; 

\path  [shading=_k3njmgcqh,_oq42pzm9h] (565.67,323.57) -- (565,230.94) -- (589.67,230.61) -- (590.33,323.24) -- cycle ; 
 \draw   (565.67,323.57) -- (565,230.94) -- (589.67,230.61) -- (590.33,323.24) -- cycle ; 

\draw    (38.67,286.33) -- (55.33,286.05) ;
\draw [shift={(58.33,286)}, rotate = 539.03] [fill={rgb, 255:red, 0; green, 0; blue, 0 }  ][line width=0.08]  [draw opacity=0] (8.93,-4.29) -- (0,0) -- (8.93,4.29) -- cycle    ;
\draw    (84.67,286.3) -- (101.33,286.02) ;
\draw [shift={(104.33,285.97)}, rotate = 539.03] [fill={rgb, 255:red, 0; green, 0; blue, 0 }  ][line width=0.08]  [draw opacity=0] (8.93,-4.29) -- (0,0) -- (8.93,4.29) -- cycle    ;
\draw    (130.67,286.3) -- (147.33,286.02) ;
\draw [shift={(150.33,285.97)}, rotate = 539.03] [fill={rgb, 255:red, 0; green, 0; blue, 0 }  ][line width=0.08]  [draw opacity=0] (8.93,-4.29) -- (0,0) -- (8.93,4.29) -- cycle    ;
\draw    (176.67,286.3) -- (193.33,286.02) ;
\draw [shift={(196.33,285.97)}, rotate = 539.03] [fill={rgb, 255:red, 0; green, 0; blue, 0 }  ][line width=0.08]  [draw opacity=0] (8.93,-4.29) -- (0,0) -- (8.93,4.29) -- cycle    ;
\draw    (221.67,286.67) -- (240.33,286.06) ;
\draw [shift={(243.33,285.97)}, rotate = 538.15] [fill={rgb, 255:red, 0; green, 0; blue, 0 }  ][line width=0.08]  [draw opacity=0] (8.93,-4.29) -- (0,0) -- (8.93,4.29) -- cycle    ;
\draw    (268.67,286.3) -- (285.33,286.02) ;
\draw [shift={(288.33,285.97)}, rotate = 539.03] [fill={rgb, 255:red, 0; green, 0; blue, 0 }  ][line width=0.08]  [draw opacity=0] (8.93,-4.29) -- (0,0) -- (8.93,4.29) -- cycle    ;
\draw    (311.67,286) -- (330.33,285.97) ;
\draw [shift={(333.33,285.97)}, rotate = 539.9100000000001] [fill={rgb, 255:red, 0; green, 0; blue, 0 }  ][line width=0.08]  [draw opacity=0] (8.93,-4.29) -- (0,0) -- (8.93,4.29) -- cycle    ;
\draw    (358.67,286.3) -- (375.33,286.02) ;
\draw [shift={(378.33,285.97)}, rotate = 539.03] [fill={rgb, 255:red, 0; green, 0; blue, 0 }  ][line width=0.08]  [draw opacity=0] (8.93,-4.29) -- (0,0) -- (8.93,4.29) -- cycle    ;
\draw    (401.67,286) -- (421.33,285.97) ;
\draw [shift={(424.33,285.97)}, rotate = 539.9200000000001] [fill={rgb, 255:red, 0; green, 0; blue, 0 }  ][line width=0.08]  [draw opacity=0] (8.93,-4.29) -- (0,0) -- (8.93,4.29) -- cycle    ;
\draw    (449.67,286) -- (470.33,285.97) ;
\draw [shift={(473.33,285.97)}, rotate = 539.9200000000001] [fill={rgb, 255:red, 0; green, 0; blue, 0 }  ][line width=0.08]  [draw opacity=0] (8.93,-4.29) -- (0,0) -- (8.93,4.29) -- cycle    ;
\draw    (499.67,286.3) -- (516.33,286.02) ;
\draw [shift={(519.33,285.97)}, rotate = 539.03] [fill={rgb, 255:red, 0; green, 0; blue, 0 }  ][line width=0.08]  [draw opacity=0] (8.93,-4.29) -- (0,0) -- (8.93,4.29) -- cycle    ;
\draw    (544.67,286.3) -- (561.33,286.02) ;
\draw [shift={(564.33,285.97)}, rotate = 539.03] [fill={rgb, 255:red, 0; green, 0; blue, 0 }  ][line width=0.08]  [draw opacity=0] (8.93,-4.29) -- (0,0) -- (8.93,4.29) -- cycle    ;
\draw    (591,286.67) -- (611.33,286.06) ;
\draw [shift={(614.33,285.97)}, rotate = 538.28] [fill={rgb, 255:red, 0; green, 0; blue, 0 }  ][line width=0.08]  [draw opacity=0] (8.93,-4.29) -- (0,0) -- (8.93,4.29) -- cycle    ;
\path  [shading=_54znauya7,_blywqi98b] (287,320.81) -- (286.34,228.18) -- (311,227.85) -- (311.66,320.48) -- cycle ; 
 \draw   (287,320.81) -- (286.34,228.18) -- (311,227.85) -- (311.66,320.48) -- cycle ; 

\path  [shading=_24tjxfdf5,_ads4frk36] (377,319) -- (376.34,226.37) -- (401,226.04) -- (401.66,318.67) -- cycle ; 
 \draw   (377,319) -- (376.34,226.37) -- (401,226.04) -- (401.66,318.67) -- cycle ; 

\draw    (68.67,225.64) .. controls (82.71,191.54) and (98.36,186.52) .. (120.94,224.61) ;
\draw [shift={(122.33,227)}, rotate = 240.16] [fill={rgb, 255:red, 0; green, 0; blue, 0 }  ][line width=0.08]  [draw opacity=0] (8.93,-4.29) -- (0,0) -- (8.93,4.29) -- cycle    ;
\draw    (159.33,227) .. controls (173.38,192.9) and (189.03,187.87) .. (211.61,225.97) ;
\draw [shift={(213,228.35)}, rotate = 240.16] [fill={rgb, 255:red, 0; green, 0; blue, 0 }  ][line width=0.08]  [draw opacity=0] (8.93,-4.29) -- (0,0) -- (8.93,4.29) -- cycle    ;
\draw    (68.67,225.64) .. controls (125.18,113.6) and (311.35,183.4) .. (340.73,224.52) ;
\draw [shift={(342.33,227)}, rotate = 240.16] [fill={rgb, 255:red, 0; green, 0; blue, 0 }  ][line width=0.08]  [draw opacity=0] (8.93,-4.29) -- (0,0) -- (8.93,4.29) -- cycle    ;
\draw    (70.33,320.33) .. controls (117.13,365.83) and (187.38,347.96) .. (210.64,324.17) ;
\draw [shift={(212.33,322.33)}, rotate = 490.86] [fill={rgb, 255:red, 0; green, 0; blue, 0 }  ][line width=0.08]  [draw opacity=0] (8.93,-4.29) -- (0,0) -- (8.93,4.29) -- cycle    ;
\draw    (115,321.67) .. controls (162.76,368.1) and (491.69,403.97) .. (577.06,324.21) ;
\draw [shift={(578.33,323)}, rotate = 495.7] [fill={rgb, 255:red, 0; green, 0; blue, 0 }  ][line width=0.08]  [draw opacity=0] (8.93,-4.29) -- (0,0) -- (8.93,4.29) -- cycle    ;
\draw    (115,321.67) .. controls (180.65,368.47) and (272.92,351.26) .. (297.88,324.42) ;
\draw [shift={(299.67,322.33)}, rotate = 488.16] [fill={rgb, 255:red, 0; green, 0; blue, 0 }  ][line width=0.08]  [draw opacity=0] (8.93,-4.29) -- (0,0) -- (8.93,4.29) -- cycle    ;
\draw    (212.33,322.33) .. controls (259.13,367.83) and (321.14,348.06) .. (343.99,324.18) ;
\draw [shift={(345.67,322.33)}, rotate = 490.86] [fill={rgb, 255:red, 0; green, 0; blue, 0 }  ][line width=0.08]  [draw opacity=0] (8.93,-4.29) -- (0,0) -- (8.93,4.29) -- cycle    ;
\draw    (257.33,226.49) .. controls (271.38,192.39) and (317.12,186.55) .. (340.9,224.61) ;
\draw [shift={(342.33,227)}, rotate = 240.16] [fill={rgb, 255:red, 0; green, 0; blue, 0 }  ][line width=0.08]  [draw opacity=0] (8.93,-4.29) -- (0,0) -- (8.93,4.29) -- cycle    ;
\draw    (257,321) .. controls (304.04,366.73) and (442.64,351) .. (483.93,325.26) ;
\draw [shift={(486.33,323.67)}, rotate = 504.94] [fill={rgb, 255:red, 0; green, 0; blue, 0 }  ][line width=0.08]  [draw opacity=0] (8.93,-4.29) -- (0,0) -- (8.93,4.29) -- cycle    ;
\draw    (257,321) .. controls (312.82,371.56) and (483.12,359.39) .. (527.72,325.24) ;
\draw [shift={(529.67,323.67)}, rotate = 499.47] [fill={rgb, 255:red, 0; green, 0; blue, 0 }  ][line width=0.08]  [draw opacity=0] (8.93,-4.29) -- (0,0) -- (8.93,4.29) -- cycle    ;
\draw    (301,227) .. controls (407.82,123.81) and (544.73,198.56) .. (572.75,228.82) ;
\draw [shift={(574.33,230.61)}, rotate = 230.31] [fill={rgb, 255:red, 0; green, 0; blue, 0 }  ][line width=0.08]  [draw opacity=0] (8.93,-4.29) -- (0,0) -- (8.93,4.29) -- cycle    ;
\draw    (301,227) .. controls (407.28,124.34) and (502.14,197.81) .. (529.06,228.35) ;
\draw [shift={(531,230.61)}, rotate = 230.31] [fill={rgb, 255:red, 0; green, 0; blue, 0 }  ][line width=0.08]  [draw opacity=0] (8.93,-4.29) -- (0,0) -- (8.93,4.29) -- cycle    ;
\draw    (299.67,322.33) .. controls (337.82,348.85) and (366.56,345.88) .. (387.73,323.16) ;
\draw [shift={(389.67,321)}, rotate = 490.86] [fill={rgb, 255:red, 0; green, 0; blue, 0 }  ][line width=0.08]  [draw opacity=0] (8.93,-4.29) -- (0,0) -- (8.93,4.29) -- cycle    ;
\draw    (351,321.67) .. controls (397.8,367.17) and (497.84,348.02) .. (527.52,325.41) ;
\draw [shift={(529.67,323.67)}, rotate = 498.81] [fill={rgb, 255:red, 0; green, 0; blue, 0 }  ][line width=0.08]  [draw opacity=0] (8.93,-4.29) -- (0,0) -- (8.93,4.29) -- cycle    ;
\draw    (383.96,227.29) .. controls (441.52,152.1) and (503.8,199.19) .. (529.13,228.4) ;
\draw [shift={(531,230.61)}, rotate = 230.31] [fill={rgb, 255:red, 0; green, 0; blue, 0 }  ][line width=0.08]  [draw opacity=0] (8.93,-4.29) -- (0,0) -- (8.93,4.29) -- cycle    ;
\draw    (437,321.67) .. controls (470.32,348.45) and (494.68,346.43) .. (527.64,325) ;
\draw [shift={(529.67,323.67)}, rotate = 506.31] [fill={rgb, 255:red, 0; green, 0; blue, 0 }  ][line width=0.08]  [draw opacity=0] (8.93,-4.29) -- (0,0) -- (8.93,4.29) -- cycle    ;

\draw (18.6,285.02) node [anchor=north west][inner sep=0.75pt]  [font=\Large,rotate=-269.44] [align=left] {Input};
\draw (620.01,297.29) node [anchor=north west][inner sep=0.75pt]  [font=\Large,rotate=-270] [align=left] {Softmax};
\draw (65.42,294.69) node [anchor=north west][inner sep=0.75pt]  [font=\Large,rotate=-270] [align=left] {Conv 3x3};
\draw (112.76,294.69) node [anchor=north west][inner sep=0.75pt]  [font=\Large,rotate=-270] [align=left] {Sep 5x5};
\draw (158.76,299.36) node [anchor=north west][inner sep=0.75pt]  [font=\Large,rotate=-270] [align=left] { Stretched};
\draw (203.42,295.39) node [anchor=north west][inner sep=0.75pt]  [font=\Large,rotate=-270] [align=left] {Conv 3x3};
\draw (250.76,295.39) node [anchor=north west][inner sep=0.75pt]  [font=\Large,rotate=-270] [align=left] {Conv 3x3};
\draw (294.09,304.39) node [anchor=north west][inner sep=0.75pt]  [font=\Large,rotate=-270] [align=left] {Dropout(p=1)};
\draw (339.42,295.39) node [anchor=north west][inner sep=0.75pt]  [font=\Large,rotate=-270] [align=left] {Avg. Pool};
\draw (382.76,296.06) node [anchor=north west][inner sep=0.75pt]  [font=\Large,rotate=-270] [align=left] {Stretched};
\draw (430.76,295.06) node [anchor=north west][inner sep=0.75pt]  [font=\Large,rotate=-270] [align=left] {Sep 3x3};
\draw (480.76,296.73) node [anchor=north west][inner sep=0.75pt]  [font=\Large,rotate=-270] [align=left] {Conv 3x3};
\draw (525.09,297.39) node [anchor=north west][inner sep=0.75pt]  [font=\Large,rotate=-270] [align=left] {Max Pool};
\draw (572.09,297.39) node [anchor=north west][inner sep=0.75pt]  [font=\Large,rotate=-270] [align=left] {Conv 3x3};

\end{tikzpicture}